\documentclass[11pt]{article}

\usepackage[final]{acl-style-files/acl}

\usepackage{times}
\usepackage{latexsym}
\usepackage[T1]{fontenc}
\usepackage[utf8]{inputenc}
\usepackage{microtype}

\usepackage{amsmath}
\usepackage{amssymb}
\usepackage{graphicx}
\usepackage{booktabs}
\usepackage{enumitem}

\newcommand{\method}{\textsc{PEEL}}
\newcommand{\ollm}{\textsc{OLLM}}
\newcommand{\contf}{Cont.\,F1}
\newcommand{\graphf}{Graph\,F1}
\newcommand{\fuzzyf}{Fuzzy\,F1}
\newcommand{\motifd}{Motif\,Dist.}
\newcommand{\litf}{Lit.\,F1}
\let\peel\method %

\title{Peeling the Matryoshka: Extracting Ontological Structure \\ from Retrieval Embeddings}

\author{
  Eunchan Kim \\
  Hanyang University \\
  \texttt{eunchan@hanyang.ac.kr} \And
  Inwhee Joe \\
  Hanyang University \\
  \texttt{iwjoe@hanyang.ac.kr}
}

\begin{document}
\maketitle

\begin{abstract}
Taxonomy construction is essential for knowledge-intensive NLP yet
remains costly: clustering-based methods (i)~need careful per-corpus
tuning and yield unstable hierarchies, while LLM-based pipelines
(ii)~depend on expensive task-specific training that does not transfer
across domains.
We report an empirical finding that bears on both: in Matryoshka
Representation Learning (MRL) embeddings, where each leading prefix is
itself a usable, coarser vector, prefix length aligns with the
coarse-to-fine granularity of gold taxonomies; shuffled- and
reverse-dimension controls isolate the alignment to the learned
early prefix dimensions. The
pattern holds across three architecturally distinct MRL models (full
statistics in \S\ref{sec:observation}).
To test whether this regularity is usable, we introduce \peel{}
(\textbf{P}eeling \textbf{E}mbeddings to \textbf{E}xtract
\textbf{L}atent ontologies): a minimal probe over these frozen
embeddings, trained once with an order-embedding objective that
represents hierarchy as vector containment; a minimum spanning
arborescence then decodes taxonomic DAGs.
Using only a concept vocabulary (no documents), \peel{} recovers
taxonomic DAGs with high structural fidelity on two \ollm{}
benchmarks, transferring to a new domain with no target-edge
supervision.
The end-to-end \ollm{} pipeline---which jointly discovers concepts
and induces structure from document corpora---is reported as a
cross-task reference (\S\ref{sec:cross-task}).
Pre-trained retrieval embeddings encode a decodable granularity
gradient; a minimal probe renders it as explicit parent--child
(is-a) structure.
\end{abstract}

\section{Introduction}
\label{sec:introduction}

Domain taxonomies, hierarchical structures of concepts linked by
taxonomic (\emph{is-a}) relations, are a fundamental form of
ontological structure and the backbone of structured knowledge
systems, from biomedical terminologies such as MeSH
\citep{lipscomb2000medical} to web-scale category hierarchies like
Wikipedia \citep{ponzetto2007deriving} and academic classification
schemes.
Throughout, \emph{ontological structure} refers to this taxonomic
(\emph{is-a}) backbone, following \ollm{}'s usage \citep{lo2024ollm}.
Yet constructing them remains one of NLP's most labor-intensive tasks.

Automated taxonomy construction has been approached from two
directions. Embedding-based methods range from flat clustering of
term vectors to recursive or geometry-based schemes (e.g.,
hyperbolic and box embeddings); the former lose hierarchy, the
latter require dedicated, hierarchy-supervised embedding spaces.
LLM-based methods instead generate hierarchy explicitly, yet demand
extensive fine-tuning or prompt engineering per domain.
Meanwhile, structural probing has shown that pre-trained
representations encode rich linguistic structure recoverable via
simple learned projections \citep{hewitt2019structural}, but has
focused on syntax, not taxonomic hierarchy.
An alternative remains unexplored: do embeddings trained purely for
retrieval already encode hierarchical signal exploitable for
taxonomy construction?

We find that they do: specifically, embeddings trained with
Matryoshka Representation Learning
\citep[MRL;][]{kusupati2022matryoshka}, whose nested prefix structure
makes truncation a principled resolution control.
Every prefix of the full vector is itself a useful representation,
concentrating coarse semantic information in early dimensions.
We observe that this coarse-to-fine organization aligns with gold
taxonomic structure: across three open-source MRL models spanning
distinct encoder families and scales (nomic, 137\,M, BERT; jina-v3,
572\,M, XLM-R; Qwen3-Embedding, 600\,M, decoder), MRL prefix
truncation consistently outperforms both a shuffled-dimension
control (ordering destroyed, content preserved) and a last-$k$
reverse control (tail dimensions), the advantage concentrated in the
learned early prefix dimensions.
A supporting ANOVA finds the label separation effect large at the
deepest taxonomic level, consistent across all model-by-dataset
sequences (Table~\ref{tab:eta}; full statistics in
\S\ref{sec:observation}), indicating the structure is not specific
to a single encoder family.

We present \peel{} (\textbf{P}eeling \textbf{E}mbeddings to
\textbf{E}xtract \textbf{L}atent ontologies), a lightweight pipeline
that exploits this signal.
Pre-trained embeddings mix taxonomic structure with retrieval-specific
features; \peel{} trains a 98K-parameter probe with an
order-embedding loss \citep{vendrov2016order} to project frozen
retrieval embeddings into a space where coordinate-wise dominance
reflects parent--child relations.
A Chu-Liu/Edmonds minimum spanning arborescence
\citep{chu1965shortest,edmonds1967optimum} then extracts a
threshold-free taxonomy tree, expanded into a DAG by greedily adding
lowest-violation edges up to a budget.

On the \ollm{} benchmarks \citep{lo2024ollm}, \peel{}---using only
the concept vocabulary, no documents---performs taxonomy structure
induction over a given concept set, recovering taxonomic DAGs with
high structural fidelity (Graph~F1 $.885 / .931$ on Wikipedia and
arXiv; cross-task comparison with the document-fed \ollm{} pipeline
in \S\ref{sec:cross-task}).

Our contributions are:
\begin{enumerate}[nosep,leftmargin=*]
  \item \textbf{Discovery.} An empirical finding that pre-trained MRL
    embeddings encode taxonomic structure in their learned early
    prefix dimensions, isolated by shuffled- and reverse-dimension
    controls and replicated across three architecturally distinct
    open-source MRL models (\S\ref{sec:observation}).
  \item \textbf{Demonstration.} \peel{}, a minimal probe decoding
    taxonomy DAGs via minimum spanning arborescence without
    task-specific embedding training (\S\ref{sec:method}).
  \item \textbf{Evaluation.} On two \ollm{} benchmarks, \peel{}
    recovers taxonomic DAGs from a concept vocabulary alone with high
    structural fidelity on the primary structure-sensitive metrics
    (Graph~F1 and \contf{}); we report the document-fed
    \ollm{} pipeline---which jointly discovers concepts \emph{and}
    induces structure from documents---as a cross-task reference
    (\S\ref{sec:experiments}).
\end{enumerate}

\section{Related Work}
\label{sec:related}

\subsection{Taxonomy Construction}

Automated taxonomy construction has evolved through several paradigms.
\emph{Pattern-based} methods, originating with Hearst patterns \citep{hearst1992automatic} and revisited at scale by \citet{roller2018hearst}, extract hypernymy from lexico-syntactic cues but suffer from low recall.
\emph{Statistical} approaches integrate heterogeneous evidence to induce taxonomies with supervised classifiers \citep{snow2006semantic} or structured prediction \citep{bansal2014structured}. Embedding-based methods such as TaxoGen \citep{zhang2018taxogen} cluster term embeddings recursively.

More recently, large language models have been applied directly.
\ollm{} \citep{lo2024ollm} fine-tunes an LLM to generate ontology subgraphs end-to-end;
Chain-of-Layer \citep{zeng2024chain} decomposes taxonomy induction into a layer-by-layer prompting procedure;
TaxoLLaMA \citep{moskvoretskii2024taxollama} unifies taxonomy tasks via instruction tuning;
and TaxoAdapt \citep{kargupta2025taxoadapt} adapts LLM-generated taxonomies to evolving corpora.

These methods share a common pattern: they either \emph{learn} taxonomy structure through task-specific training or \emph{generate} it via prompted LLMs.
LLMs4OL \citep{giglou2024llms4ol} organises ontology learning into three tasks---term typing, taxonomy discovery, and non-taxonomic relation extraction---and \peel{} targets taxonomy discovery: recovering \emph{is-a} pairs over a given set of types.
Whether pre-trained embeddings---trained solely for retrieval---already encode exploitable taxonomic structure has remained unexplored.

\subsection{Hierarchical Representation Learning}

Order Embeddings \citep{vendrov2016order} encode hierarchy directly in embedding geometry, modelling asymmetric \emph{is-a} relations via coordinate-wise inequalities: a parent's coordinates are no larger than its child's in every dimension. Later work extends this ordering to density-based formulations \citep{athiwaratkun2018hierarchy}.
Poincar\'{e} Embeddings \citep{nickel2017poincare} use hyperbolic space to capture tree-like hierarchies.

Most closely related to our approach, HiTs \citep{he2024hits} retrains pre-trained sentence transformers in Poincar\'{e} space with a centripetal loss to produce a hierarchy encoder.
While both HiTs and \peel{} map pre-trained language models to hierarchical structure, they differ fundamentally:
HiTs \emph{injects} hierarchy by retraining the entire encoder (22M--109M parameters) with a hierarchy-aware objective,
whereas \peel{} \emph{discovers} hierarchy latent in frozen retrieval embeddings via a 98K-parameter probe.

\subsection{Structural Probing of Pre-trained Representations}

What linguistic structure pre-trained representations encode has been studied extensively through probing classifiers \citep{belinkov2019analysis, conneau2018cram, tenney2019bert}.
Most directly relevant, \citet{hewitt2019structural} demonstrate that a linear transformation of BERT representations recovers parse tree distances, establishing the \emph{structural probe} paradigm: a simple learned projection can extract complex structure from frozen representations.
Subsequent work has examined what probing accuracy actually reveals: \citet{hewitt2019control} propose control tasks to measure probe selectivity, and \citet{voita2020mdl} frame probing as compression, arguing that a probe's minimum description length---not just accuracy---determines whether structure is genuinely encoded.

Crucially, existing structural probes target \emph{syntactic} structure (dependency and constituency trees).
\peel{} extends this paradigm to \emph{taxonomic} structure: we show that a learned projection of retrieval embeddings, trained with an order embedding loss, recovers parent--child relations in concept hierarchies---a qualitatively different structural signal from syntax.

\subsection{Matryoshka Representation Learning}

Matryoshka Representation Learning \citep[MRL;][]{kusupati2022matryoshka} trains embeddings so that every prefix of the full vector is itself a valid representation, producing a nested family of $O(\log d)$ representations from a single forward pass.
Originally designed for efficient retrieval at variable compute budgets, MRL models are now widely deployed in production systems. Its authors already noted that this nesting ``naturally captures the underlying hierarchy of the class labels'' in a two-level ImageNet superclass setting (their \S5).

\citet{hanley2025matryoshka} instead engineer it: they train custom Matryoshka embeddings with a modified contrastive loss that explicitly encodes news similarity hierarchies into different dimension ranges, an approach we term \emph{hierarchy injection}.
We establish the complementary result: standard MRL models---trained only for retrieval, with no hierarchical supervision---already exhibit prefix structure that tracks depth in a multi-level gold concept taxonomy, an alignment that \S\ref{sec:observation} isolates with shuffled- and reverse-dimension controls and replicates across three architecturally distinct encoders.
This \emph{hierarchy discovery} requires no custom training and applies to any MRL-capable encoder.

\paragraph{Summary.}
\peel{} sits at the intersection of four threads: it applies the structural probe paradigm (\S2.3) with order embedding geometry (\S2.2) to frozen MRL representations (\S2.4) for taxonomy construction (\S2.1), requiring neither task-specific embedding training nor LLM generation.

\section{MRL Prefixes Encode Taxonomic Structure}
\label{sec:observation}

As noted in \S\ref{sec:introduction}, MRL embeddings concentrate
coarse semantic information in early dimensions by design
\citep{kusupati2022matryoshka}.
We hypothesize that this coarseness aligns with \emph{taxonomic
structure}: shorter prefixes capture broad category distinctions
while longer prefixes preserve finer-grained structure.

\paragraph{Experimental design.}
We evaluate kNN label purity ($k{=}10$) across three MRL models
spanning distinct encoder families and scales: Nomic-embed-text-v1.5
\citep{nussbaum2025nomic} (137\,M, BERT), jina-embeddings-v3
\citep{sturua2024jina} (572\,M, XLM-R), and Qwen3-Embedding
\citep{zhang2025qwen3embedding} (600\,M, decoder), on two
taxonomic benchmarks.
We compare MRL prefix truncation against two controls:
(i)~a \emph{shuffled} control that randomly permutes all dimensions
before truncation, preserving total information content while
destroying the learned dimension ordering
\citep[cf.\ control tasks;][]{hewitt2019control}, and
(ii)~a \emph{reverse} control that truncates the last $k$ dimensions,
excluding the front-loaded learned structure.

\begin{figure*}[t]
\centering
\includegraphics[width=\textwidth]{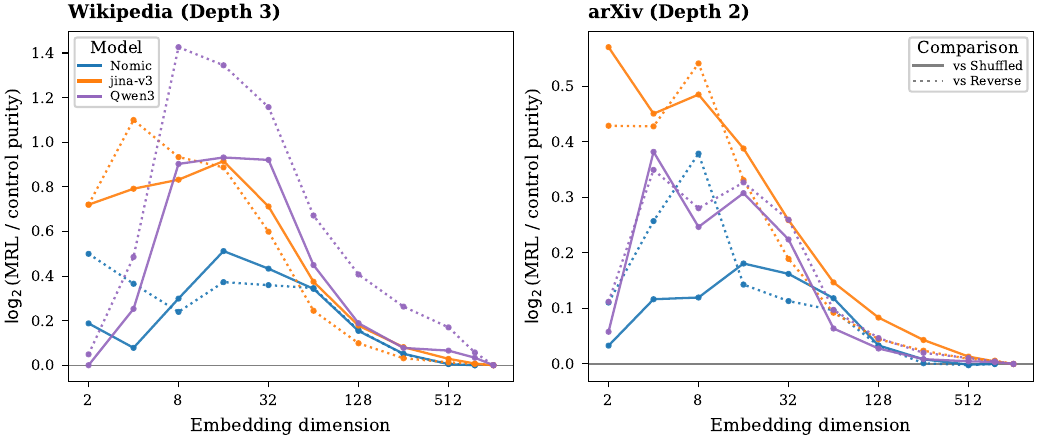}
\caption{Relative advantage of MRL prefix truncation over two
controls, by embedding dimension, on Wikipedia and arXiv at the
deepest gold taxonomic level. $y = \log_2(\text{MRL} \div
\text{control kNN purity})$; $y{=}0$ is parity with the control.
Solid $=$ vs.\ shuffled control (dimension ordering destroyed);
dotted $=$ vs.\ reverse control (last-$k$ tail dimensions). Colors
denote three architecturally distinct MRL models (Nomic, jina-v3,
Qwen3). All curves rise above parity and peak within the early
dimensions; at full dimensionality the prefix equals the complete
embedding, so the controls coincide by construction.}
\label{fig:knn-gradient}
\end{figure*}

\paragraph{Results.}
MRL prefix truncation outperforms \emph{both} controls (the
order-destroying shuffled control and the last-$k$ tail-dimension
reverse control) across all three architecturally distinct encoder
families, with the advantage concentrated in the early, truncated
prefix dimensions (Figure~\ref{fig:knn-gradient}).
At full dimensionality the prefix coincides with the complete
embedding, so all three coincide by construction; the separation
therefore reflects the \emph{learned early-prefix ordering}.
That the advantage holds for every model against \emph{both} a
randomly permuted and a tail-truncated control isolates the signal to
this learned ordering.
This directional consistency holds for every model (one-sided sign
test; Appendix~\ref{app:signtest}).
As supporting triangulation, we quantify how strongly taxonomic
labels separate at each depth via one-way ANOVA over every embedding
dimension, reporting the mean $\eta^2$ (Table~\ref{tab:eta}):
on Wikipedia, effects are small-to-medium at Depth~1 (.04--.07),
medium at Depth~2 (.13--.18), and Cohen's-large at Depth~3
(.34--.40); the shallower arXiv taxonomy (two gold levels) shows the
same direction (Depth~1 .04--.08, Depth~2 .12--.20).
The depth ordering of effect sizes holds without exception across all
six model$\times$dataset sequences.
This $\eta^2$ comparison serves only as a supporting cross-model
consistency check.

\begin{table}[t]
\centering
\small
\setlength{\tabcolsep}{3pt}
\renewcommand{\arraystretch}{0.92}
\begin{tabular}{ll ccc}
\toprule
\textbf{Model} & \textbf{Data} & \textbf{Depth\,1} & \textbf{Depth\,2} & \textbf{Depth\,3} \\
\midrule
Nomic (137M)   & wiki  & .058 & .159 & .364 \\
jina-v3 (572M) & wiki  & .067 & .181 & .400 \\
Qwen3 (600M)   & wiki  & .043 & .131 & .337 \\
\midrule
Nomic (137M)   & arXiv & .039 & .122 & --- \\
jina-v3 (572M) & arXiv & .078 & .205 & --- \\
Qwen3 (600M)   & arXiv & .039 & .131 & --- \\
\bottomrule
\end{tabular}
\caption{ANOVA effect sizes ($\eta^2$) for kNN label purity by
taxonomy depth, reported as a supporting cross-model consistency
check. arXiv has max depth~2. The depth ordering holds in every row.
Because gold class counts differ sharply across depths (Wikipedia:
33/875/7124 at Depth~1/2/3), these values index consistency,
\emph{not} an absolute cross-depth magnitude. Cohen's benchmarks:
$.01$ small, $.06$ medium, $.14$ large \citep{cohen1988statistical}.}
\label{tab:eta}
\end{table}

The signal's concentration in early dimensions, with higher
dimensions adding little over the controls, suggests that full
embeddings mix taxonomic structure with retrieval-specific features
that dilute hierarchy.
A learned projection trained to preserve parent--child ordering
should therefore separate these components.
A training-free pre-probe check is reported in
Appendix~\ref{app:preprobe}.
\S\ref{sec:method} operationalizes this insight.

\section{Method}
\label{sec:method}

\begin{figure*}[t]
\centering
\includegraphics[width=0.82\textwidth]{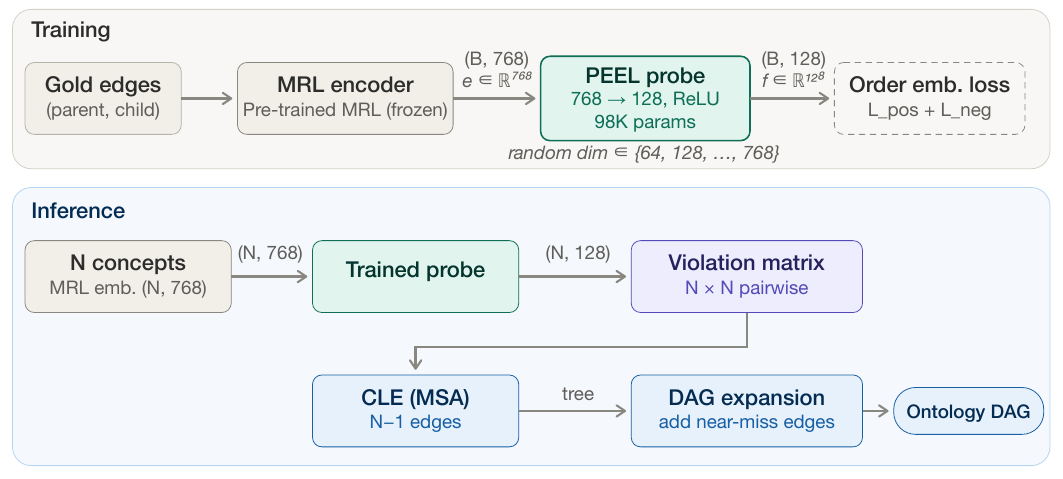}
\caption{The \peel{} pipeline. \textbf{Training} (top): frozen MRL embeddings are projected through a probe (98K parameters) trained with order embedding loss under random dimensional truncation. \textbf{Inference} (bottom): the trained probe maps $N$ concept embeddings to a 128-d order-embedding space; a pairwise violation matrix is constructed and resolved by Chu-Liu/Edmonds' algorithm into a threshold-free tree, then expanded into a directed acyclic graph via near-miss edges.}
\label{fig:pipeline}
\end{figure*}

\peel{} operationalizes the observation from
\S\ref{sec:observation}---that MRL prefixes encode taxonomic
structure---through three stages (Figure~\ref{fig:pipeline}):
(1)~projecting frozen embeddings into an order-embedding space
    \citep{vendrov2016order} via a probe trained once on
    source-domain gold edges,
(2)~extracting a threshold-free taxonomic tree via Chu-Liu/Edmonds
    minimum spanning arborescence
    \citep{chu1965shortest,edmonds1967optimum}, and
(3)~expanding the tree into a DAG by adding low-violation edges.
Each component is deliberately simple: \S\ref{sec:observation}
establishes that the hierarchical signal is already present in
pre-trained embeddings; \peel{} demonstrates that it is
operationally exploitable with simple machinery.

\subsection{Order-Embedding Probe}
\label{sec:probe}

Pre-trained MRL embeddings mix taxonomic signal with
retrieval-specific features in overlapping dimensions
(\S\ref{sec:observation}).
Following the structural probing paradigm
\citep{hewitt2019structural}, we train a learned projection to
separate these components:
\begin{equation}
  \mathbf{f} = \operatorname{ReLU}(W\mathbf{e} + \mathbf{b}),
  \quad W \in \mathbb{R}^{d_\text{out} \times D}
  \label{eq:projection}
\end{equation}
We take this paradigm as conceptual inspiration---a minimal probe
can surface latent structure---and \peel{} serves as a demonstration
that the structure \S\ref{sec:observation} establishes is
operationally exploitable, not as a diagnostic claim about the
embeddings.
ReLU projects outputs into the non-negative orthant
$\mathbb{R}_+^{d_\text{out}}$ required by the order-embedding
framework \citep{vendrov2016order}.
We set $d_\text{out}{=}128$ based on ablation
(Table~\ref{tab:dout-sweep}).

We train the projection using order-embedding loss
\citep{vendrov2016order}:
\begin{align}
  \mathcal{L} &= \mathcal{L}_+ + \mathcal{L}_-
  \label{eq:oeloss} \\[3pt]
  \mathcal{L}_+ &= \tfrac{1}{|\mathcal{P}|}
    {\textstyle\sum_{(p,c) \in \mathcal{P}}}
    \bigl\|\max(\mathbf{0},\,
          \mathbf{f}_p {-} \mathbf{f}_c)\bigr\|^2
  \label{eq:oe-pos} \\[3pt]
  \mathcal{L}_- &= \tfrac{1}{|\mathcal{N}|}
    {\textstyle\sum_{(p,n) \in \mathcal{N}}}
    \max\!\bigl(0,\; m \notag \\
    &\quad {-} \bigl\|\max(\mathbf{0},\,
          \mathbf{f}_p {-} \mathbf{f}_n)\bigr\|^2\bigr)
  \label{eq:oe-neg}
\end{align}
where $\mathcal{L}_+$ encourages $\mathbf{f}_p \leq \mathbf{f}_c$
for positive parent--child pairs~$\mathcal{P}$, and $\mathcal{L}_-$
pushes negatives apart by margin~$m$ ($\mathcal{N}$~=~negative pairs).
Because \S\ref{sec:observation} isolates this signal using controls
that involve no probe training, the learned projection recovers a
granularity ordering the embeddings already encode rather than
imposing it through supervision.

We choose order embeddings for two reasons:
(1)~they directly encode the asymmetric is-a relation, unlike
    symmetric contrastive losses; and
(2)~the coordinate-wise ordering constraint yields the violation
    score used for tree extraction (\S\ref{sec:cle}).

\subsection{Random Truncation Training}
\label{sec:rtt}

Datasets differ in domain and structural profile, so no single
truncation captures the full taxonomic structure.
We exploit MRL's prefix-truncation property: at each training step,
we randomly sample $d \in \{64, 128, 256, 384, 512, 768\}$,
truncate all embeddings to $d$~dimensions, and zero-pad back to
$D{=}768$ before projection.
This produces a single model that handles all resolutions;
at inference, we select the best dimensionality per dataset.
Training at a single truncation length would tie the probe to that
resolution's feature distribution; random truncation instead forces
it to learn order-embedding relationships that hold across all
resolutions simultaneously (Table~\ref{tab:ablation}).

\subsection{Tree Extraction via Chu-Liu/Edmonds}
\label{sec:cle}

Given projected embeddings~$\mathbf{f}$, we define the violation
score between nodes $i$ and $j$ as:
\begin{equation}
  v(i,j) = \sum_{k=1}^{d_\text{out}}
    \bigl[\max(0,\, f_k(i) - f_k(j))\bigr]^2
  \label{eq:violation}
\end{equation}
A low $v(i,j)$ indicates that $\mathbf{f}_i \leq \mathbf{f}_j$
coordinate-wise, suggesting $i$ is a parent of~$j$.

We construct a directed graph weighted by violation scores and
extract the minimum spanning arborescence (MSA) via the
Chu-Liu/Edmonds algorithm
\citep{chu1965shortest,edmonds1967optimum},
producing an optimal rooted tree with exactly
$N{-}1$ edges.
CLE requires no threshold, unlike containment-ratio or
edge-weight thresholding approaches.

\subsection{DAG Expansion}
\label{sec:dag}

A tree contains exactly $N{-}1$ edges, which may undercount the
true is-a relations in a polyhierarchical ontology.
To recover additional edges, we rank all non-tree node pairs by
violation score and greedily add the lowest-violation pairs while
maintaining acyclicity, up to an edge budget
$n_\text{expand}$ (\S\ref{sec:setup}).

\section{Experiments}
\label{sec:experiments}

\subsection{Setup}
\label{sec:setup}

\paragraph{Datasets.}
We evaluate on two datasets from the \ollm{} benchmarks \citep{lo2024ollm}, which pair document corpora with gold taxonomies.
The two datasets span distinct domains and structural profiles.
Wiki-ol provides 8{,}033 concept terms with a gold taxonomy of 14{,}673 edges at depth~3.
ArXiv-ol contains 61 categories in a shallow hierarchy of depth~2.

\paragraph{Evaluation Metrics.}
We report five metrics from the \ollm{} evaluation suite \citep{lo2024ollm}: \litf{} (literal edge overlap, \ollm{}'s sanity-check reference), \fuzzyf{} (thresholded edge matching; all-MiniLM-L6-v2, $\tau{=}0.436$), \contf{} (threshold-free Hungarian edge matching, scored by the weaker of the two endpoint similarities), \graphf{} (Hungarian node matching over SGConv neighbourhood embeddings), and \motifd{} (triadic-census total-variation divergence).
We designate \graphf{} and \contf{} as the primary structure-sensitive metrics: they capture global structural similarity and edge-endpoint correspondence, and are robust to surface-level label variation.

\paragraph{Baselines.}
All baselines are from \citet{lo2024ollm} with reported numbers: Memorisation (training edges verbatim), Hearst Patterns, Zero-shot and Finetune (Mistral-7B), and the full \ollm{} pipeline.
\peel{} receives only the gold concept vocabulary; all LLM-based baselines (Zero-shot, Finetune, \ollm{}) additionally consume each article's title and abstract per the \ollm{} protocol (\citealp{lo2024ollm}, Figs.~5--6), and Hearst extracts from the source corpus.
For arXiv-ol, \ollm{} (transfer) fine-tunes on 2{,}048 additional arXiv samples; \peel{} uses no arXiv training data (zero-shot).

\paragraph{Implementation.}
\peel{} uses Nomic-embed-text-v1.5 \citep{nussbaum2025nomic} as the frozen encoder (768-dimensional).
The probe is a single linear layer (768$\to$128) followed by ReLU, totaling 98K trainable parameters.
We train with Order Embedding loss (margin $m{=}4.0$) using Adam (learning rate $5{\times}10^{-4}$, batch size 512) on wiki-ol gold parent--child edges for 10 epochs on a single GPU (${\sim}4$\,seconds).\footnote{Code and data: \url{https://github.com/Eunchan24/PEEL}}

\paragraph{Training and inference protocol.}
\label{sec:hp-protocol}
The probe is trained on wiki-ol gold parent--child edges, with the training graph split 80\%/10\%/10\% into train/validation/holdout (edge-level split); each epoch we track validation loss and edge-ranking AUROC, retaining the lowest-loss checkpoint over the 10-epoch schedule.
Negative pairs use a $10{:}1$ negative-to-positive ratio, mixing $50\%$ random nodes, $30\%$ sibling distractors (other children of the same parent), and $20\%$ reversed pairs.
For tree extraction, the virtual root connects to every node at edge weight equal to the $95$th percentile of positive violation scores.
Wiki-ol inference hyperparameters (prefix truncation $d{=}512$, DAG expansion budget $n_\text{expand}{=}15{,}000$, CLE candidate count $k{=}10$) are chosen on wiki-ol validation performance; Table~\ref{tab:ablation} reports per-$d$ sensitivity.
For arxiv-ol, the wiki-ol-trained probe is applied unchanged, with inference settings ($d{=}768$, $k{=}20$, tree-only) following from the dataset's compact 61-concept structure rather than from arxiv-ol-specific tuning; no arxiv-ol data is used for training, validation, or hyperparameter selection.

\begin{table*}[t]
\centering
\footnotesize
\renewcommand{\arraystretch}{0.9}
\begin{tabular}{ll l ccccc}
\toprule
\textbf{Dataset} & \textbf{Method} & \textbf{Input} & \textbf{Lit.F1}$\uparrow$ & \textbf{Fuz.F1}$\uparrow$ & \textbf{Cont.F1}$\uparrow$ & \textbf{Grph.F1}$\uparrow$ & \textbf{Motif.D}$\downarrow$ \\
\midrule
\multicolumn{8}{l}{\textit{Concept-only methods (no document access)}} \\
\midrule
wiki-ol & Memorisation     & training edges    & .134 & .837          & .314          & .419          & .063 \\
        & PEEL (ours)      & concepts only     & .111          & .877          & .551 & .885 & .000 \\
\midrule
\multicolumn{8}{l}{\textit{Document-using methods (corpus + concepts)}} \\
\midrule
wiki-ol & Hearst Patterns  & corpus + concepts & .003          & .538          & .350          & .544          & .163 \\
        & Zero-shot (M-7B) & title + abstract  & .007          & .871          & .455          & .639          & .341 \\
        & Finetune (M-7B)  & title + abstract  & .124          & .884          & .470          & .588          & .050 \\
        & OLLM             & title + abstract  & .093          & .915 & .500          & .644          & .080 \\
\midrule
arXiv-ol & OLLM (transfer)  & title + abstract & .040 & .570 & .357 & .633 & .097 \\
         & PEEL (zero-shot) & concepts only    & .298 & .748 & .604 & .931 & .011 \\
\bottomrule
\end{tabular}
\caption{Main results on \ollm{} benchmarks, grouped by input modality. LLM-based baselines additionally consume each article's title and abstract per the \ollm{} protocol (\citealp{lo2024ollm}, Figs.~5--6); \ollm{} (transfer) fine-tunes Mistral-7B on 2{,}048 additional arXiv samples. \peel{} rows use a gold concept vocabulary and a probe trained on wiki-ol gold edges (\S\ref{sec:setup}), the arXiv-ol row zero-shot; \ollm{} instead discovers concepts and structure from documents, so the \ollm{} scores are a cross-task reference, not a ranking. \motifd{} values below $10^{-4}$ are reported as $.000$.}
\label{tab:main-results}
\end{table*}

\subsection{Structural Fidelity from a Concept Vocabulary}
\label{sec:main-results}

\peel{} (a 98K-parameter probe receiving \emph{only} the concept vocabulary) recovers taxonomic DAGs with \graphf{} $.885$ on wiki-ol (supervised, edge-level 80/10/10 split) and $.931$ on arXiv-ol (zero-shot transfer from the wiki-ol-trained probe). \ollm{}, which consumes the document corpus and fine-tunes Mistral-7B to jointly discover concepts and induce structure, scores $.644$ and $.633$ on the same benchmarks as a cross-task reference (\S\ref{sec:cross-task}).
Table~\ref{tab:main-results} reports all five metrics.

The probe trained on $80\%$ of wiki-ol gold edges (\S\ref{sec:hp-protocol}) recovers $.885$ \graphf{} across the full $8{,}033$-concept structure, generalizing the taxonomic signal beyond the training-edge-replay floor (Memorisation: $.419$).

\paragraph{Wiki-ol.}
On wiki-ol's 8{,}033 concepts, \contf{} rises from $.500$ (\ollm{}) to \peel{}'s $.551$.
On the two label-string metrics (\litf{}, \fuzzyf{}), LLM label-generation systems hold an expected advantage; Memorisation tops \litf{} ($.134$) by replaying training edges verbatim, and \ollm{} leads \fuzzyf{} ($.915$) via title-string overlap.

\paragraph{ArXiv-ol (zero-shot transfer).}
\label{sec:arxiv-results}
The wiki-ol-trained probe transfers zero-shot to arXiv-ol's 61 categories: \peel{} attains \graphf{} $.931$ and \contf{} $.604$; \ollm{}, which additionally fine-tunes Mistral-7B on 2{,}048 arXiv samples, scores $.633$ and $.357$.

\subsection{Cross-Task Reference}
\label{sec:cross-task}

\peel{} performs structure induction from a concept vocabulary alone.
\ollm{} addresses a broader task---joint concept discovery \emph{and} structure induction---and, like the Mistral-7B baselines, consumes article text.
The comparison is therefore asymmetric in both task scope and input: \peel{} is given strictly less.
We report it for reference, not as a same-conditions claim.

\subsection{Ablation Studies}
\label{sec:ablation}

\paragraph{Embedding dimension.}
Table~\ref{tab:ablation} varies the input prefix truncation $d$ with probe output fixed at $d_\text{out}{=}128$.
\contf{} peaks at $d{=}512$ ($.551$) and drops at full dimensionality ($.479$ at $d{=}768$), matching \S\ref{sec:observation}'s coarse-to-fine prefix gradient.
We select $d{=}512$ on validation \contf{}; \graphf{} stays in $.879$--$.906$ across dimensions, peaking at $d{=}768$.
Even with $d_\text{out}{=}32$ (24K parameters; Appendix~\ref{sec:appendix-ablation}, Table~\ref{tab:dout-sweep}), \peel{} yields \graphf{} $.870$ and \contf{} $.517$.

\begin{figure*}[t]
\centering
\includegraphics[width=\textwidth]{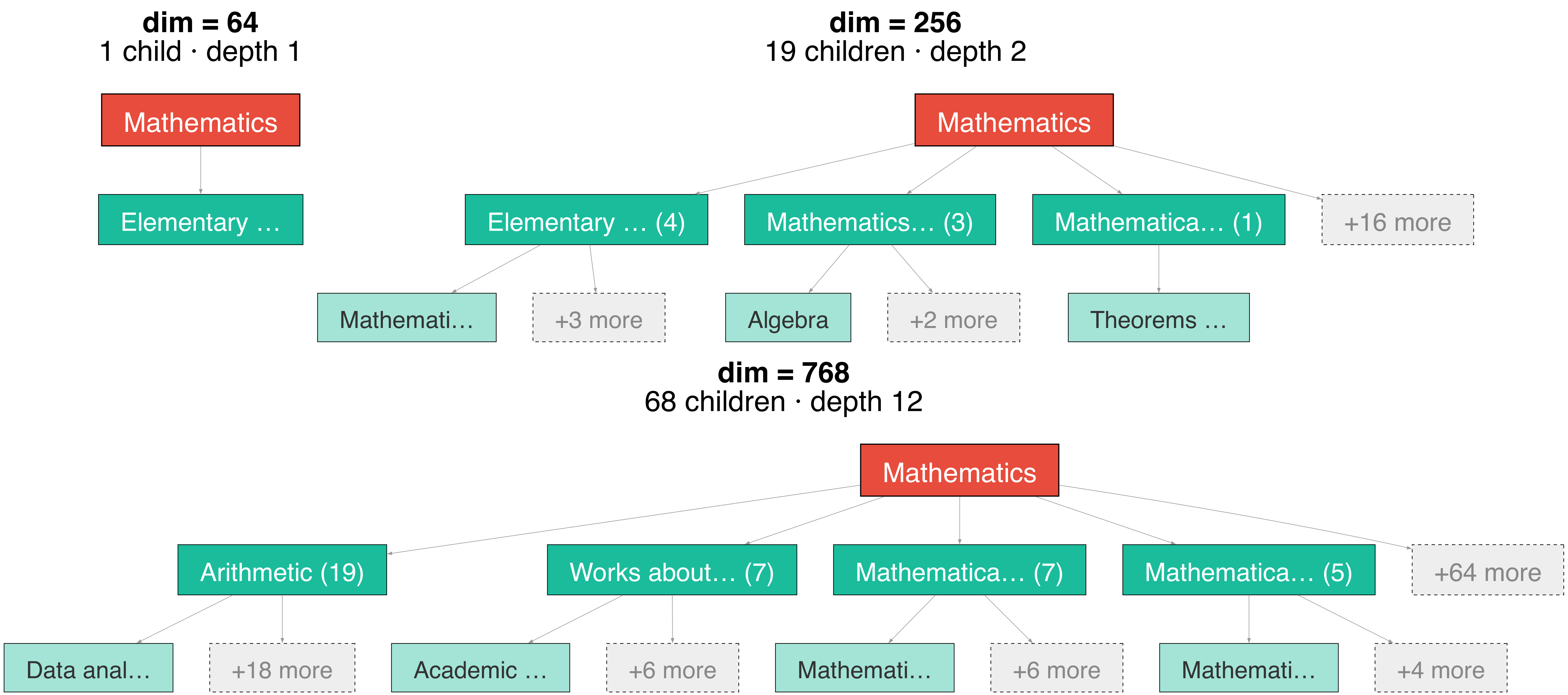}
\caption{Mathematics subtree extracted at three prefix dimensions
from the trained probe: at $d{=}64$, only one child is recovered;
at $d{=}256$, 19~children emerge across depth~2; at $d{=}768$,
68~children span 12~levels of depth. Increasing prefix dimension
progressively unfolds fine-grained hierarchical structure.}
\label{fig:mathematics-subtree}
\end{figure*}

\paragraph{Multi-resolution decomposition.}
We extract Chu-Liu/Edmonds trees at each truncation point.
Edge-set Jaccard between all dimension pairs stays below $.18$, with adjacent dimensions ($d{=}512$ vs.\ $d{=}768$: $.178$) sharing more edges than distant ones ($d{=}64$ vs.\ $d{=}768$: $.018$).
Across all five dimensions, $99\%$ of nodes change parents: prefix truncation yields \emph{non-nested decompositions}, each prefix length inducing a qualitatively distinct hierarchy across resolutions.
At the category level, the effect is vivid: the \emph{Mathematics} subtree grows from 1~child at $d{=}64$ to 19~at $d{=}256$ and 68~at $d{=}768$, illustrating how increasing prefix dimension progressively unfolds fine-grained taxonomic structure (Figure~\ref{fig:mathematics-subtree}).

\section{Discussion}
\label{sec:discussion}

\paragraph{What the probe reveals about embedding structure.}
A 98K-parameter probe recovers taxonomic DAGs (Graph~F1
$.885$ on Wikipedia, $.931$ on arXiv) from \emph{frozen} retrieval
embeddings, showing that the prefix-ordering regularity
\S\ref{sec:observation} isolates is decodable into explicit
parent--child structure without retraining the encoder.
The coarse-to-fine gradient isolated by shuffled- and
reverse-dimension controls (\S\ref{sec:observation}) implies that
MRL's nested-prefix training implicitly organises semantic
granularity along the dimension axis, a structural regularity
that, to our knowledge, has not been previously documented in
standard MRL encoders trained purely for retrieval.

\paragraph{Euclidean order embeddings vs.\ hyperbolic alternatives.}
\peel{}'s Euclidean order-embedding probe avoids the
encoder-retraining required by hyperbolic alternatives such as
HiTs \citep{he2024hits}, achieving structural fidelity with no
hierarchy-aware encoder training (geometry comparison in
Appendix~\ref{app:hyperbolic}).

\paragraph{Each prefix length captures a different hierarchy.}
\S\ref{sec:ablation} reports that CLE trees at different prefix
lengths yield non-nested, complementary decompositions.
This multi-resolution property---extracting multiple taxonomic
granularities from a single forward pass over a frozen retrieval
embedding, with no retraining or per-resolution recomputation---directly
leverages MRL's nested-prefix design.
The kNN purity gradient (\S\ref{sec:observation}) and the
prefix-level tree disjointness are complementary: the former is a
local neighbourhood measure, the latter a global parent--child
topology, so prefixes can share neighbourhoods yet decode distinct
global hierarchies.

\paragraph{Structure induction as a modular component.}
\peel{} performs structure induction \emph{from a concept
vocabulary alone}, isolating it as a separable subproblem: a
98K-parameter probe over frozen embeddings recovers taxonomic DAGs
at Graph~F1 $.885$ and $.931$ with no document access. Any
concept-discovery front-end---an LLM extractor, a corpus miner, or
a curated vocabulary---can supply that vocabulary.
\ollm{}, which couples concept discovery and structure induction
in a single document-fed pipeline, addresses a broader task and
serves as a cross-task reference (\S\ref{sec:cross-task}).
Pairing \peel{} with any such front-end yields an end-to-end
taxonomy pipeline without retraining the structure-induction stage.

\section{Conclusion}
\label{sec:conclusion}

Pre-trained Matryoshka Representation Learning (MRL) embeddings
organise semantic granularity along the dimension axis; shuffled-
and reverse-dimension controls isolate this gradient to the learned
early prefix dimensions across three architecturally distinct MRL
models (\S\ref{sec:observation}).
Exploiting this signal, \peel{} (a 98K-parameter probe over
\emph{frozen} embeddings, applied to a concept vocabulary
alone) recovers taxonomic DAGs with high structural fidelity
(Graph~F1 $.885$ on Wikipedia, $.931$ on arXiv); the document-fed
\ollm{} pipeline is reported as a cross-task reference
(\S\ref{sec:cross-task}).

A retrieval-efficiency objective thus yields a multi-resolution
taxonomic geometry that \peel{} can decode.

Integrating \peel{} with corpus-based concept discovery for
end-to-end taxonomy construction is a natural direction for
future work.

\section*{Limitations}

Our work has the following limitations.
(1)~Our evaluation covers two domains from the \ollm{} benchmarks
\citep{lo2024ollm} spanning distinct structural profiles (deep
polyhierarchy and shallow tree).
\peel{} transfers zero-shot from wiki-ol to arXiv-ol
(\S\ref{sec:arxiv-results}).
Our validated transfer covers target ontologies no deeper than the
source domain; how far this reach extends is genuinely open.
Extension to specialised domains
such as biomedicine and law is a natural direction for future work.
(2)~We adopt the five structural metrics established by the
\ollm{} benchmark \citep{lo2024ollm} (Lit~F1, Fuzzy~F1, Cont~F1,
Graph~F1, Motif Distance), which evaluate taxonomy structure
directly.
Label realisation quality and downstream utility (e.g.,
retrieval, question answering) are outside this paper's scope.
(3)~The resolution-control analysis in \S\ref{sec:observation}
relies on MRL's nested prefix-truncation property.
The probe itself (\S\ref{sec:method}) operates on any
fixed-dimensional embedding.
(4)~\peel{} assumes a gold concept vocabulary and source-domain
training edges; transfer requires no target-domain edges.

\section*{Ethics Statement}

This work extracts taxonomic structure from publicly available
pre-trained embedding models and evaluates on published benchmarks
\citep{lo2024ollm}.
No human subjects, private data, or crowd-sourced annotations are
involved.
The taxonomies we construct reflect biases present in the
pre-training corpora of the underlying MRL models; users deploying
\method{} in sensitive domains should audit the resulting
hierarchies for undesirable biases or omissions.
Our method's low computational cost (seconds on a single GPU)
reduces the environmental footprint relative to approaches that
require fine-tuning large language models.

\bibliography{references}

\appendix

\section{Additional Ablation Studies}
\label{sec:appendix-ablation}

\begin{figure}[t]
\centering
\includegraphics[width=\columnwidth]{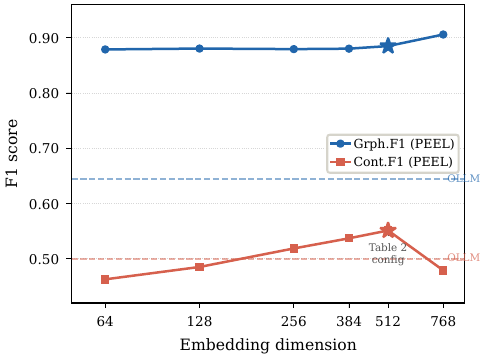}
\caption{PEEL performance by embedding dimension on wiki-ol ($d_\text{out}{=}128$ fixed). $^\star$ marks the Table~\ref{tab:main-results} configuration (dim=512). Dashed lines show OLLM baselines.}
\label{fig:dim-performance}
\end{figure}

Table~\ref{tab:ablation}, referenced from \S\ref{sec:method} and
\S\ref{sec:experiments}, reports \peel{}'s primary ablation---input
prefix truncation $d$.
Figure~\ref{fig:dim-performance} visualises this sweep, showing
\peel{}'s \graphf{} and \contf{} relative to the \ollm{} reference
baselines across the tested dimensions.

\begin{table}[t]
\centering
\footnotesize
\setlength{\tabcolsep}{4pt}
\begin{tabular}{l cc}
\toprule
\textbf{Configuration} & \textbf{Cont.F1}$\uparrow$ & \textbf{Grph.F1}$\uparrow$ \\
\midrule
\multicolumn{3}{l}{\textit{Input prefix truncation $d$ ($d_\text{out}{=}128$ fixed)}} \\
\midrule
\quad $d = 64$   & .463 & .879 \\
\quad $d = 128$  & .485 & .880 \\
\quad $d = 256$  & .519 & .879 \\
\quad $d = 384$  & .537 & .880 \\
\quad $d = 512$\rlap{$^\star$}  & .551 & .885 \\
\quad $d = 768$  & .479 & .906 \\
\midrule
\quad \ollm{} (reference) & .500 & .644 \\
\bottomrule
\end{tabular}
\caption{Input prefix truncation $d$ ablation on wiki-ol. $^\star$ denotes the Table~\ref{tab:main-results} configuration. \ollm{} (reference) included for the same metrics.}
\label{tab:ablation}
\end{table}

\paragraph{Probe output dimension.}
Table~\ref{tab:dout-sweep} varies $d_\text{out}$ with input
dimension fixed at $d{=}512$.
Even at $d_\text{out}{=}32$ (24K parameters), \peel{} achieves
\graphf{} $.870$ and \contf{} $.517$.
Performance improves monotonically with $d_\text{out}$, confirming
that additional projection capacity helps but is not strictly
necessary.

\begin{table}[t]
\centering
\small
\begin{tabular}{l r cc}
\toprule
$d_\text{out}$ & \textbf{Params} & \textbf{\contf{}} $\uparrow$ & \textbf{\graphf{}} $\uparrow$ \\
\midrule
32  & 24K  & .517 & .870 \\
64  & 49K  & .544 & .878 \\
128\rlap{$^\star$} & 98K  & .551 & .885 \\
\midrule
\ollm{} (ref) & 7B & .500 & .644 \\
\bottomrule
\end{tabular}
\caption{Probe output dimension sweep on wiki-ol ($d{=}512$ fixed).
$^\star$ denotes the Table~\ref{tab:main-results} configuration.}
\label{tab:dout-sweep}
\end{table}

\paragraph{Edge budget sensitivity.}
Table~\ref{tab:nexpand} varies the DAG expansion budget
$n_\text{expand}$ with all other hyperparameters fixed.
\graphf{} remains above $.868$ across all tested budgets
(10K--25K).
\contf{} peaks at $n_\text{expand}{=}15{,}000$, the configuration
used in Table~\ref{tab:main-results}.
The stability across budgets confirms that $n_\text{expand}$ is
not a sensitive hyperparameter; Table~\ref{tab:edgepr} reports the
corresponding literal edge precision and recall.

\begin{table}[t]
\centering
\small
\begin{tabular}{r cc c}
\toprule
$n_\text{expand}$ & \textbf{\contf{}} $\uparrow$ & \textbf{\graphf{}} $\uparrow$ & \textbf{\motifd{}} $\downarrow$ \\
\midrule
10,000  & .503 & .899 & .000 \\
12,000  & .533 & .893 & .000 \\
15,000\rlap{$^\star$} & .551 & .885 & .000 \\
20,000  & .500 & .876 & .001 \\
25,000  & .451 & .868 & .001 \\
\midrule
\ollm{} (ref) & .500 & .644 & .080 \\
\bottomrule
\end{tabular}
\caption{DAG expansion budget ($n_\text{expand}$) sensitivity on
wiki-ol. $^\star$ denotes the Table~\ref{tab:main-results}
configuration. \graphf{} is stable across all tested budgets.}
\label{tab:nexpand}
\end{table}

\begin{table}[t]
\centering
\small
\begin{tabular}{r cc}
\toprule
$n_\text{expand}$ & \textbf{Lit-P} $\uparrow$ & \textbf{Lit-R} $\uparrow$ \\
\midrule
tree only & 16.9 & 9.3 \\
10,000  & 14.6 & 10.0 \\
12,000  & 12.8 & 10.5 \\
15,000\rlap{$^\star$} & 11.0 & 11.2 \\
20,000  & 9.2 & 12.6 \\
25,000  & 8.0 & 13.7 \\
\bottomrule
\end{tabular}
\caption{Literal edge precision and recall by expansion budget on
wiki-ol (\%). $^\star$ denotes the Table~\ref{tab:main-results}
configuration; the reported setting is balanced rather than
recall-maximizing. See \S\ref{sec:setup} for the primary
structure-sensitive metrics.}
\label{tab:edgepr}
\end{table}

\section{Qualitative Analysis}
\label{sec:appendix-qualitative}

\begin{figure*}[t]
\centering
\includegraphics[width=\textwidth]{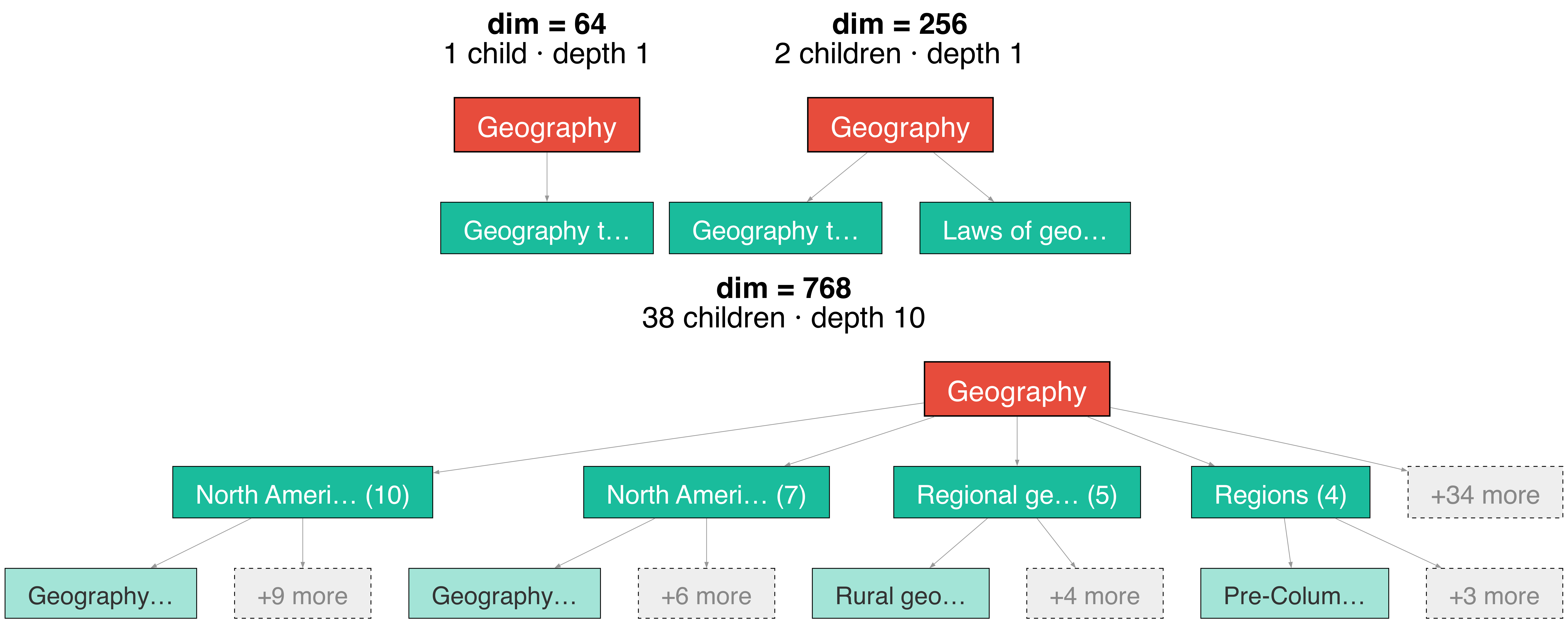}
\caption{\emph{Geography} subtree at three prefix dimensions. At
$d{=}64$, only 1~child is recovered at depth~1; at $d{=}256$,
2~children remain at depth~1; at $d{=}768$, 38~children span
depth~10. Compare with Mathematics
(Figure~\ref{fig:mathematics-subtree}).}
\label{fig:zoom-geography}
\end{figure*}

\begin{figure*}[t]
\centering
\includegraphics[width=\textwidth]{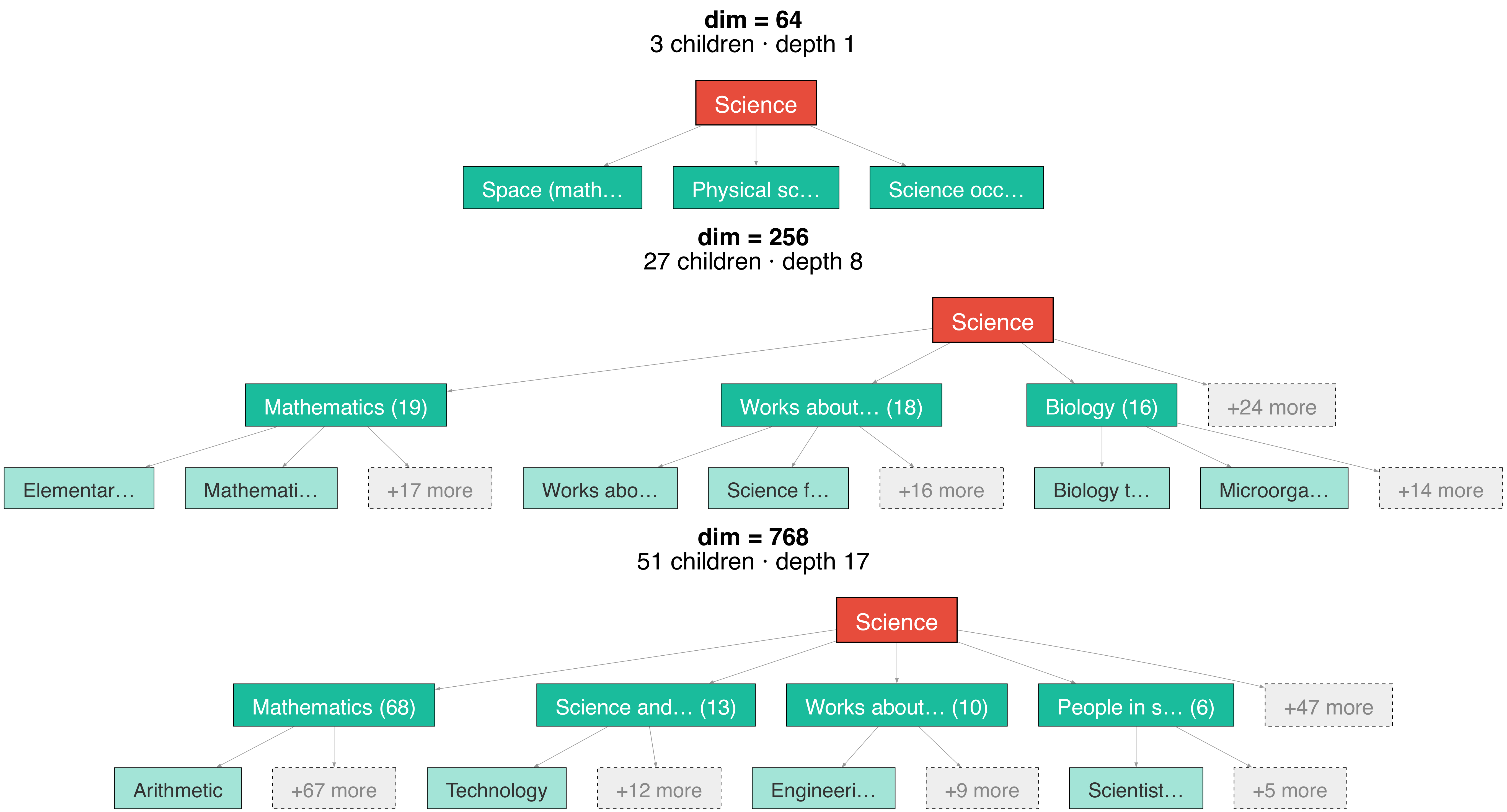}
\caption{\emph{Science} subtree at three prefix dimensions. At
$d{=}64$, 3~children appear at depth~1; at $d{=}256$, 27~children
span depth~8; at $d{=}768$, 51~children span depth~17. Together
with Geography (Figure~\ref{fig:zoom-geography}) and Mathematics
(Figure~\ref{fig:mathematics-subtree}), these examples confirm
that prefix truncation induces qualitatively distinct hierarchical
decompositions across diverse categories, consistent with the
low inter-dimension Jaccard similarity ($< .18$) reported in
\S\ref{sec:ablation}.}
\label{fig:zoom-science}
\end{figure*}

\paragraph{Dimensional sensitivity of tree topology.}
CLE trees extracted at different dimensions exhibit structurally
distinct topologies (e.g., 10 top-level categories at $d{=}64$
vs.\ 17 at $d{=}512$ on arXiv-ol), consistent with the
dimensional sensitivity observed in
Figure~\ref{fig:knn-gradient}.
Figures~\ref{fig:zoom-geography} and~\ref{fig:zoom-science}
extend the \emph{Mathematics} example from
Figure~\ref{fig:mathematics-subtree} to two additional categories,
confirming that prefix-dependent unfolding of fine-grained
structure generalizes across diverse subtrees.

\paragraph{Edge overlap vs.\ structural similarity.}
On arXiv-ol, \peel{} reproduces 18 of 60 gold edges exactly
(30\%), while achieving \graphf{} $.931$.
This gap reflects the difference between exact edge matching and
SGConv-based structural similarity: \peel{} captures global
topology even when individual parent--child assignments differ
from the gold standard.

\section{Hyperbolic Comparison: Order Embeddings vs.\ Poincar\'{e} Approaches}
\label{app:hyperbolic}

\peel{}'s probe operates in Euclidean space with coordinate-wise
dominance constraints \citep{vendrov2016order}, rather than in
hyperbolic space \citep{nickel2017poincare}.
This appendix details the comparison referenced in
\S\ref{sec:discussion}.

\paragraph{Hyperbolic encoders.}
Poincar\'{e}-based approaches such as HiTs \citep{he2024hits}
retrain the entire encoder ($22$M--$109$M parameters) with
hierarchy-aware losses, exploiting the exponential volume growth
of hyperbolic space to capture tree-like hierarchies with low
distortion.

\paragraph{Euclidean order embeddings.}
\peel{} trains only a $98$K-parameter projection on top of
frozen MRL embeddings using an order-embedding loss, requiring no
hierarchy-aware pre-training of the encoder.
The coordinate-wise dominance constraint directly encodes the
asymmetric is-a relation, and the violation score derived from
this constraint provides the edge weight for CLE tree extraction
(\S\ref{sec:cle}).

\paragraph{Comparison summary.}
Both approaches map pre-trained language models to hierarchical
structure, but differ on two axes: (i)~\emph{trainable
parameters}---$22$M--$109$M (HiTs) versus $98$K (\peel{}), a
three-order-of-magnitude reduction; (ii)~\emph{training
objective}---hierarchy-aware contrastive loss applied to the
entire encoder (HiTs) versus order-embedding loss applied only to
a lightweight probe over frozen embeddings (\peel{}).
A direct head-to-head benchmark comparison between HiTs and
\peel{} on the \ollm{} taxonomies is left for future work.

\section{Directional-Consistency Counts}
\label{app:signtest}

Table~\ref{tab:signtest} reports, for each model and benchmark, the
number of prefix$\times$depth conditions in which MRL prefix
truncation exceeds the shuffled-dimension control, with the
combined-benchmark one-sided sign test (null $p_0{=}0.5$). Every per-model
$p$ is below $10^{-6}$. The four non-wins in the Nomic--arXiv cell
are the full-dimensionality conditions, where the prefix coincides
with the complete embedding and MRL, shuffled, and reverse are
identical by construction; the cell remains significant
($16/20$, $p{=}5.9{\times}10^{-3}$).

\begin{table}[t]
\centering
\small
\begin{tabular}{l ccc c}
\toprule
\textbf{Model} & \textbf{Wiki} & \textbf{arXiv} & \textbf{Comb.} & \textbf{one-sided $p$} \\
\midrule
Nomic   & 26/30 & 16/20 & 42/50 & $5.8{\times}10^{-7}$ \\
jina-v3 & 30/33 & 20/22 & 50/55 & $1.1{\times}10^{-10}$ \\
Qwen3   & 29/33 & 20/22 & 49/55 & $9.1{\times}10^{-10}$ \\
\bottomrule
\end{tabular}
\caption{Per-condition MRL\,$>$\,shuffled win counts
(prefix\,$\times$\,depth) and combined one-sided sign test, by model
and benchmark. All per-model $p<10^{-6}$.}
\label{tab:signtest}
\end{table}

\section{Raw Geometry Before the Probe}
\label{app:preprobe}

Before any probe training, the frozen embeddings already recover
which of two concepts is more general: a training-free measure over
raw prefixes scores $69.1\%$ on gold parent--child pairs and
$91.3\%$ on ancestor--descendant pairs at depth gap ${\ge}2$, while
equal-granularity siblings sit at chance ($50.0\%$).
Explicit, coordinate-aligned \emph{is-a} containment remains the
probe's contribution: order violation on the raw axes is at chance
under a random-rotation null.

\end{document}